\pdfoutput=1

\documentclass[11pt]{article}

\usepackage{ACL2023}

\usepackage{times}
\usepackage{latexsym}

\usepackage[T1]{fontenc}

\usepackage[utf8]{inputenc}

\usepackage{microtype}

\usepackage{inconsolata}

\usepackage{amsmath}
\usepackage{hyperref}
\usepackage{url}
\usepackage{booktabs}
\usepackage{graphicx}
\usepackage{verbatim}

\newcommand{\eg}{e.g., }

\AtBeginDocument{}%
\AtBeginDocument{}%

\title{Pragmatic Inference with a CLIP Listener \\ for Contrastive Captioning}

\author{Jiefu Ou$^1$ \quad Benno Krojer$^2$ \quad Daniel Fried$^1$ \\
         Carnegie Mellon University$^1$ \quad Mila/McGill University$^2$ \\
        {\tt jiefuo@andrew.cmu.edu}}

\begin{document}
\maketitle
\begin{abstract}
We propose a simple yet effective and robust method for contrastive captioning: generating discriminative captions that distinguish target images from very similar alternative distractor images. Our approach is built on a pragmatic inference procedure that formulates captioning as a reference game between a \emph{speaker}, which produces possible captions describing the target, and a \emph{listener}, which selects the target given the caption. 
Unlike previous methods that derive both speaker and listener distributions from a single captioning model, we leverage an off-the-shelf CLIP model to parameterize the listener. 
Compared with captioner-only pragmatic models, our method benefits from rich vision-language alignment representations from CLIP when reasoning over distractors. 
Like previous methods for discriminative captioning, our method uses a hyperparameter to control the tradeoff between the informativity (how likely captions are to allow a human listener to discriminate the target image) and the fluency of the captions. However, we find that our method is substantially more robust to the value of this hyperparameter than past methods, which allows us to automatically optimize the captions for informativity --- outperforming past methods for discriminative captioning by 11\% to 15\% accuracy in human evaluations.\footnote{The code is available at \url{https://github.com/JefferyO/prag_clip_contra_caption}}
\end{abstract}

\section{Introduction}
\begin{figure}[t]
\centering
    \includegraphics[width=1.0\linewidth]{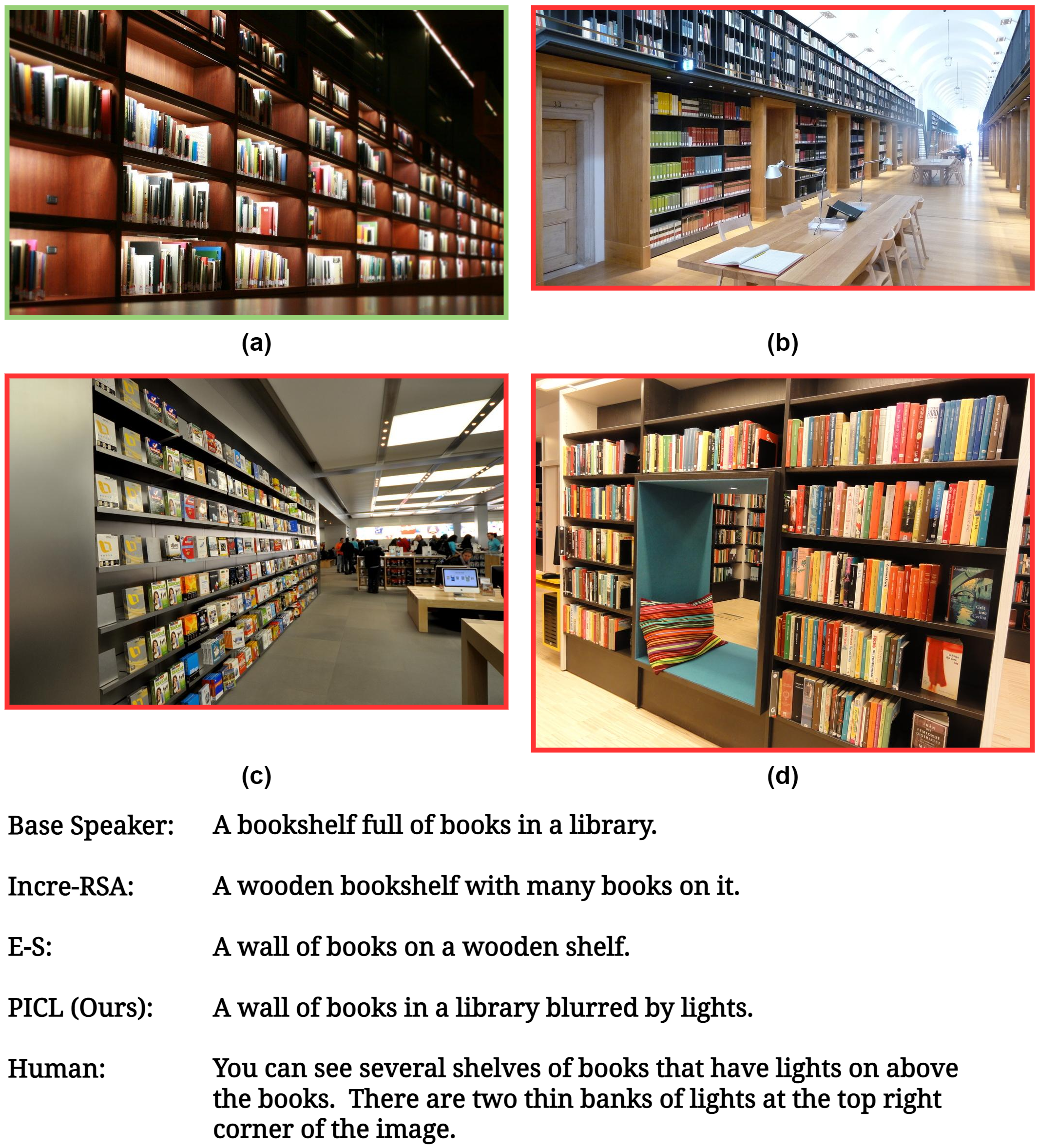}
    \caption{Illustration of the contrastive captioning task with a random example from the ImageCoDe dataset. Models are tasked with generating captions that distinguish the target image (a) from other very similar distractors images (b) to (d). (There are a total of 9 distractors in each set of images, we omit the rest of them for simplicity of illustration.) Compared with baselines from previous work, our proposed approach, PICL, generates informative captions that help clearly identify the target out of the distractors, while remaining natural and fluent.}
    \label{fig:intro}
    \vspace{-1em}
\end{figure}
Discriminative captioning provides a challenging testbed for generating context-sensitive grounded language. In this task, a model must produce a description of a \emph{target image} (\eg the green highlighted image in \autoref{fig:intro}) that allows a person to correctly identify the target image from among a set of similar \emph{distractor images} (\eg the red highlighted images).
Good captions must strike a balance between two criteria: (1) being fluent descriptions of the target image and (2) being discriminative in context: allowing a person to pick out the target image from the set.

Past work on discriminative captioning has successfully applied techniques from computational pragmatics to trade off between the two criteria above~\citep{andreas-klein-2016-reasoning,Vedantam-etal-2017-ContextAwareCF,cohn-gordon-etal-2018-pragmatically}. Possible captions are selected using a combination of two scoring functions: (1) the caption's probability under a standard image captioning model, or \emph{base speaker score}, which measures the caption's fluency and faithfulness to the image, and (2) a \emph{base listener score}, which predicts how likely a human listener would be to correctly identify the target image given the caption, i.e. measuring discriminativeness.These past works typically obtain the listener scores from the image captioning (speaker) model itself, for example using Bayesian inference over the set of possible images~\citep{cohn-gordon-etal-2018-pragmatically}.
The relative weight of these two scores is controlled using a \emph{informativity hyperparameter},\footnote{This parameter is also sometimes referred to as a \emph{rationality parameter}.} whose value affects the tradeoff between producing captions that are predicted to be fluent and faithful, versus captions that are predicted to be discriminative.
It is challenging to automatically choose a value for this hyperparameter, as captions that appear to be discriminative under a captioning model are frequently uninformative for people~\cite{dessi-etal-2022-communication}.

Our approach, \textbf{PICL} (\textbf{P}ragmatic \textbf{I}nference with a \textbf{C}LIP \textbf{L}istener) follows this same pragmatic framework, but scores discriminativeness using a listener model separate from the speaker. We implement the listener model using CLIP~\cite{Radford-etal-2021-learningtv}. As shown in previous work, the rich vision-language representation learned in CLIP (1) provides robust assessments of model-generated captions that highly correlate with human judgments~\cite{hessel-etal-2021-clipscore}, and (2) effectively quantifies the degree of discriminativeness/informativeness of visual referring expressions~\cite{takmaz-etal-2022-less}.

To evaluate PICL, we conduct experiments with sets of images from ImageCoDe \cite{krojer-etal-2022-image}, a challenging dataset originally designed for contrastive retrieval: retrieving target images among a set of distractors given contextual descriptions. We perform contrastive captioning on this dataset for the first time. We compare PICL to past work on two criteria: (1) informativeness and (2) fluency, evaluating both metrics using automatic as well as human evaluations.

Results show that our approach typically outperforms past methods on both criteria, and is substantially more robust to the value of the informativity hyperparameter.
In particular, we are able to choose this hyperparameter automatically by maximizing how informative the captions are predicted to be to human evaluators. In contrast, we find that maximizing predicted informativity leads past methods to produce captions that are so disfluent that they are misleading for people. In this automatic hyperparameter selection setting, our method produces captions that are 11\% to 15\% easier  for human annotators to interpret correctly than past work.

\section{Related Work}
\label{sec:related work}

\paragraph{Contrastive Captioning}
A variety of methods for contrastive captioning generate captions that optimize for discriminative objectives, e.g., minimizing the textual similarity between captions for the target and distractor images \cite{wang2020compare}, using generated captions as input to image retrieval models \cite{Luo2018DiscriminabilityOF, li2018show}, and computing CLIP similarity scores between captions and target images \cite{cho-etal-2022-fine}. Other methods involve leveraging fine-grained image regional features to generate distinctive captions based on similar and/or unique objects among target and distractors \cite{Wang2021GroupbasedDI, Mao2022RethinkingTR}, paraphrasing generic captions to enhance both diversity and informativeness \cite{Liu2019GeneratingDA}, and finetuning RL-optimized caption models to encourage low-frequency words \cite{Honda2022SwitchingTD}.
Most of the methods above require training a discriminative captioning model --- either by designing an discriminative captioning architecture that takes multiple images as input, or fine-tuning a model using discriminative rewards.  In contrast, our proposed approach is \textit{fully inference-time} --- it requires no training, and is applicable to any off-the-shelf generic captioning model. 

Our approach builds on a family of inference-time pragmatic-based contrastive captioning methods which have taken one of two approaches: (1) incrementally generating captions but using only a captioning model (our \emph{speaker} model), where tokens are chosen that have high probability for the target image and low probability for the distractor~\cite{Vedantam-etal-2017-ContextAwareCF,cohn-gordon-etal-2018-pragmatically,nie-etal-2020-pragmatic} or (2) using a separate discriminative model
but selecting a discriminative caption from among a set of entire captions generated by the speaker model for the target image~\cite{andreas-klein-2016-reasoning,luo2017comprehension}. 
Our work shows that these approaches can be productively combined, using a strong off-the-shelf discriminative model (CLIP) to guide the incremental generation of captions.
This allows us to tackle a more challenging dataset and task than previous discriminative captioning work, containing a large number (10) of highly-similar distractor images.

\paragraph{Pragmatics} Our approach to contrastive generation follows a long line of work on computational pragmatics, particularly in the Rational Speech Acts framework~\cite{frank2012predicting,goodman2016pragmatic} which models language generation as an interaction between speakers and listeners. Prior work has found that pragmatic generation can improve performance on a variety of NLP tasks, including reference games \citep{monroe-etal-2017-colors}, instruction generation~\citep{fried-etal-2018-unified}, summarization~\citep{shen-etal-2019-pragmatically}, machine translation~\citep{cohn-gordon-goodman-2019-lost}, and dialogue~\citep{kim-etal-2020-will,fried-etal-2021-reference}. 

\paragraph{Tradeoff between discriminativeness and accuracy/fluency}
Assessing the quality of image captions requires multifaceted evaluation. Prior work on contrastive/discriminative captioning investigates the tradeoff of model performance between discriminativeness and accuracy/fluency \cite{Wang2021GroupbasedDI, Liu2019GeneratingDA, Honda2022SwitchingTD, cho-etal-2022-fine, Vedantam-etal-2017-ContextAwareCF, andreas-klein-2016-reasoning}. In this paper, we also perform an extensive study on the tradeoff between informativeness and fluency. Specifically, we focus on analyzing the robustness of the proposed and baseline methods in the tradeoff according to the selection of hyperparameters.

\section{Method}
\label{sec:method}

Our PICL approach conducts incremental pragmatic inference at the token level by combining a base speaker and a CLIP listener to derive a pragmatic speaker. At each step of decoding, the base speaker selects a set of candidate tokens and adds them to partial captions. Given candidate partial captions, the listener updates its beliefs on which is the target among the set of images based on CLIP similarity measurement. In particular, it contrasts each partial caption to all the images by calculating the CLIP similarity scores of partial caption-image pairs and normalizes over all images to derive the listener likelihood. Finally, a pragmatic speaker reasons over both the base speaker and listener by combining their distribution to rerank partial captions, select a highly-scored subset and proceed to the next decoding step.

\subsection{Incremental Pragmatic Inference Framework}
Similar to \newcite{cohn-gordon-etal-2018-pragmatically}, we formulate the process of generating contrastive captions as a series of reference games between two agents, a \emph{speaker} and a \emph{listener}. Given a shared visual context $\mathcal{I} = i^+ \cup \mathcal{I^-}$ consisting of a target image $i^+$ and a set of $m$ similar distractors $\mathcal{I^-} = \{i^{-}_{1}, \dots, i^-_m \}$, the speaker aims to produce a sequence of $T$ tokens $o_{1:T} = (o_1, \dots o_T)$ that could let the listener identify $i$ from $I$. Such pragmatic inference is conducted \textit{incrementally}: at each step $t$ of the caption generation, the speaker selects the next token $o_t$ by playing the reference game with the listener based on the context $I$ and the partial caption $o_{<t}$ obtained from the last step. In the following subsections, we will introduce the speaker and listener models as well as the incremental inference strategy in detail.

\subsection{Speaker and Listener Models}
\label{subsec:method_models}
\textbf{Base Speaker} \quad At each step of generation, the \textit{base speaker} $S_0$ yields a distribution $P_{S_0} (o_t | o_{<t}, i^+)$ over the token vocabulary for the next possible token $o_t$, conditioning on the previous partial caption and the target image. We parameterize $P_{S_0}$ with a context-agnostic captioning model. In particular, we use OFA\footnote{We use the OFA-base configuration from \url{https://github.com/OFA-Sys/OFA}} \cite{Wang-etal-2022-unifyingat}, a unified sequence-to-sequence multimodal pretraining model and finetune it on MSCOCO Image Captioning dataset \cite{chen-etal-2015-microsoftcc}. Finetuned OFA is a strong base captioner; at the time of this work, it achieves state-of-the-art performance on MSCOCO Image Captioning. \\

\noindent\textbf{Base Listener} \quad Given a candidate partial caption $o_{1:t} = (o_{<t}, o_t)$ generated by $S_0$, the base listener $L_0$ yields a distribution $P_{L_0}(i | o_{1:t}, \mathcal{I})$ over all candidate images $i \in \mathcal{I}$, modeling the likelihood of choosing each candidate given the partial caption at step $t$ and the shared context $\mathcal{I}$. We derive $P_{L_0}$ from a zero-shot CLIP model by normalizing its similarities between images and partial captions over all image candidates:
\begin{align}
    P_{L_0}(i | o_{1:t}, \mathcal{I}) = \frac{\exp(c(i, o_{1:t}))}{\Sigma_{i^{'} \in \mathcal{I}}\exp(c(i^{'}, o_{1:t}))}
\end{align}
where $c(i, o_{1:t})$ denotes the cosine similarity between the CLIP visual encoding of $i$ and textual encoding of $o_{1:t}$\\

\noindent\textbf{Pragmatic Speaker} \quad From the base speaker and listener, we derive a distribution for the pragmatic speaker $S_1$ as 
\begin{align}
    P_{S_1}(o_t | o_{<t}, i^+, \mathcal{I}) = &P_{L_0}(i^+ | o_{1:t}, \mathcal{I})^{\lambda} \nonumber \\ \cdot &P_{S_0} (o_t | o_{<t}, i^+) ^ {1 - \lambda} \label{eq2}
\end{align}
where $\lambda \in [0, 1]$ is a ``informativity'' hyper-parameter that trades off between producing fluent (from $S_0$) and informative 
 (from $L_0$) captions.

\subsection{Decoding with Approximation}
\label{subsec:decode}
To iteratively generate captions with the pragmatic speaker $S_1$, we perform beam search with beam width $B$, which involves solving
\begin{align}
    \arg \max_{o_t} P_{S_1}(o_t | o_{<t}, i^+, \mathcal{I}) \label{eq3}
\end{align}
for each beam item. However, it is computationally infeasible to obtain the exact solution to \autoref{eq3} since it requires encoding all \#(vocabulary size) possible next partial captions with CLIP to calculate $P_{L_0}$ at each step. Thus, we adopt a sub-sampling approach similar to \newcite{andreas-klein-2016-reasoning, fried-etal-2018-unified}. At each step of decoding, a subset of $N (N > B)$ candidate next partial captions $o_{1:T}$ are obtained via beam search from the base speaker distribution $P_{S_0}$, and these $N$ candidates are rescored with \autoref{eq2} to approximate \autoref{eq3}. Finally, only the top $B$ candidates after rescoring are retained to continue with.

\section{Experimental Setup}
\label{sec:exp}
We evaluate PICL on ImageCoDe~\cite{krojer-etal-2022-image}, a dataset originally designed for image retrieval with contextual descriptions. Given the high visual similarity of the images in each problem in the dataset, we adopt it as a challenging testbed for discriminative captioning. We evaluate PICL and competitive baseline methods on two criteria, informativeness and fluency, using both automatic and human evaluation. For informativeness,
we follow previous work~\cite{cohn-gordon-etal-2018-pragmatically,newman-etal-2020-communication} to automatically evaluate the performance of pragmatic models with an \textit{evaluating listener} $L_{eval}$. The discriminativeness of the method being evaluated is quantified by the retrieval accuracy of $L_{eval}$ with method-generated captions as input. For fluency, we score the well-formedness of generated captions with the perplexity (PPL) under GPT-2~\citep{Radford-etal-2019-LanguageMA}.

In addition to the automatic evaluation, we conduct human evaluation where annotators are tasked to a) retrieve the target image given the caption and b) score the fluency of the caption.

\subsection{Dataset}
We use sets of images collected in ImageCoDe \cite{krojer-etal-2022-image} to evaluate the proposed approach. Each image set in ImageCoDe consists of 10 visually similar images.
The image sets are collected in two categories: \textit{static pictures} and \textit{video frames}.
A random subset of images per set is selected as targets, for which human annotators write discriminative captions that are retained if other humans can successfully use it to retrieve the target.

In our experiments, we use the validation split of ImageCoDe for hyper-parameter selection and evaluate model performance on the test split. 
The valid and test sets contain 1,039 and 1,046 sets of images and 2,302 and 2,306 human written captions, respectively. 

\autoref{tab:imagecode-acc-human-caption} shows the retrieval performance of several models on ImageCoDe test split, where \textbf{CLIP-zero-shot} is the base listener used in PICL and \textbf{ALBEF-finetuned} is the evaluating listener used for automatic evaluation (see \autoref{subsec:auto_eval}). Given the large performance gap of all models between static and video subsets, we believe the video frames are too challenging for current neural models to make pragmatic and contextual inferences for both captioning and retrieving. Therefore, we use only static images in our experiments.

\subsection{Automatic Evaluation}
\label{subsec:auto_eval}
\noindent \textbf{Informativeness} \quad Following \citet{cohn-gordon-etal-2018-pragmatically} and \citet{newman-etal-2020-communication}, we evaluate the informativeness of captions generated by our method and baselines using a \textit{listener test}: whether an \emph{evaluative listener} model could identify the target image out of the distractors, given generated captions.
However, an evaluative listener can only be an imperfect proxy for human listeners, and past work has found that utterances that are informative to an evaluative listener model can be uninterpretable to people, a phenomenon known as codebooking~\cite{kim-etal-2019-codraw} or language drift~\cite{lazaridou-etal-2020-multi}. This issue is particularly likely to complicate evaluation in a pragmatic framework like ours, where an explicit listener model (a frozen CLIP model, in our PICL approach) is used to guide utterance generation. 

To mitigate this codebooking issue in evaluation, past work has made the evaluative listener dissimilar by training it on separate data~\cite{cohn-gordon-etal-2018-pragmatically,kim-etal-2019-codraw,fried-etal-2021-reference}; we additionally use a separate architecture for the evaluative listener, dissimilar from our CLIP listener: the ALBEF vision-language model~\cite{Li-etal-2021-alignbf}.
We finetune ALBEF on the human-written contextual captions for the retrieval task in ImageCode.\footnote{Specifically, we finetuned the refcoco-checkpoint contrastively, i.e. with the 9 distractors in the same batch.} As shown in \autoref{tab:imagecode-acc-human-caption}, finetuned ALBEF outperforms the best-performing retrieval model from previous work \cite{krojer-etal-2022-image} on ImageCoDe with human-written captions, so we use ALBEF-finetuned as our evaluating listener in automatic evaluations of informativeness. 
\begin{table}[t]
\centering
\begin{tabular}{lccc}
\toprule
& \multicolumn{1}{c}{All} & \multicolumn{1}{c}{Video} & \multicolumn{1}{c}{Static} \\ \midrule
CLIP-zero-shot      & 22.4 & 15.6 &  47.8 \\
CLIP-finetuned-best      & 29.9 & 22.0 & 59.8 \\
ALBEF-finetuned     & 33.6 & 22.7 & 74.2 \\
\bottomrule
\end{tabular}
\caption{Retrieval accuracy on ImageCoDe test split with human-written contextual captions as input. In the proposed method, we use CLIP-zero-shot as the base listener and ALBEF-finetuned as the listener for evaluation. CLIP-finetuned denotes the best-performing model in previous work. The fine-tuned ALBEF outperforms the best CLIP model with a large margin on static images while improving slightly on video frames. Comparing with performances on static images, all models struggle on video frames.
}
\vspace{-1em}
\label{tab:imagecode-acc-human-caption}
\end{table}

\noindent\textbf{Fluency} \quad While being informative, discriminative captions should also be natural and fluent. Therefore, we additionally perform automatic evaluations of the fluency of generated captions by computing their perplexity using a GPT-2 language model~\cite{Radford-etal-2019-LanguageMA}.

\subsection{Human Evaluation}
\label{subsec: human_eval}
Recent analysis on ImageCode \cite{dessi-etal-2022-communication} and in other reference game settings \cite{lazaridou-etal-2020-multi} reveals that utterances generated by neural models can be discriminative enough for other neural models to retrieve the target image while being misleading to humans. This implies that the performance of a neural retriever evaluative listener (e.g., ALBEF) on model-generated captions might not correctly reflect the degree of informativeness of the captions from a human's perspective. Therefore, we further conduct a human evaluation for PICL and baseline methods on Amazon MTurk, where we present human workers with the same image retrieval task as for ALBEF, and use the success rate of workers in identifying the correct target images (\textbf{retrieval accuracy}) to measure the informativeness of the given captions. 
To obtain human judgments of caption fluency, we additionally ask workers to score the captions on a Likert scale ranging from 1 (nonsense) to 5 (completely natural).
We randomly sampled 100 sets of static images from the ImageCoDe test split and select one image with the human-written caption as the target. For each target, we produce a caption with each model and, together with the original human caption, present each caption-set pair to 3 workers. More details about the human evaluation setup could be found in \autoref{subsec:appendix_hit}.

\subsection{Baselines}
We compare PICL to three baselines: 

\noindent\textbf{Base Speaker} \quad We use the base speaker $S_0$ introduced in \autoref{sec:method}. The base speaker takes only the target image as input and generates context-agnostic captions regardless of the distractors. 

\noindent\textbf{Incre-RSA} \quad We further implement the incremental RSA model (Incre-RSA) from \citet{cohn-gordon-etal-2018-pragmatically} as a competitive baseline. Specifically, we derive the Bayesian RSA model introduced in \newcite{cohn-gordon-etal-2018-pragmatically} from our base speaker $S_0$, which enables direct comparison with our proposed approach. Unlike PICL, Incre-RSA does not have a separate model as the listener. The listener probabilities are derived with Bayesian inference at each decoding step based on the speaker distribution and an image prior.

\noindent\textbf{E-S} \quad Also based on $S_0$, we implement the \textit{emitter-suppressor} (E-S) beam search introduced in \citet{Vedantam-etal-2017-ContextAwareCF} for discriminative image captioning.  Similar to Incre-RSA, the E-S approach differs from PICL mainly in that it does not contain a separate model to rescore partial captions from a listener's perspective. Instead, it incorporates contextual reasoning by selecting tokens that, under the base speaker, have high probability for the target image but low probability for the distractor images, using a weighted difference of scores. 
Since their task and model formulation considers only a single distractor image, we extend it to include all distractors in the set by calculating the suppressor distribution as the mean of the distribution of the next token conditioned on each of the distractors.

For all three baselines, we use beam search at inference with the same beam width $B$ as PICL. 

\subsection{Informativity Hyperparameter Selection}
\label{subsec:hparam_select}
Both our PICL method and the Incre-RSA and E-S baselines use an informativity hyperparameter\footnote{Sometimes also referred to as a ``rationality'' parameter.} to trade off between predicted informativity and fluency in generated captions.
We describe two methods for choosing a value for this hyperparameter for each method.

\paragraph{Informativity Maximization}
In our primary set of experiments, we set the informativity hyperparameter for each method automatically to maximize the performance of our evaluating listener, ALBEF, on the captions in the validation set.
We refer to the models obtained under this scheme as \textbf{PICL}, \textbf{Incre-RSA}, and \textbf{E-S}, respectively.

\begin{figure}[t]
\centering
    \includegraphics[width=1.0\linewidth]{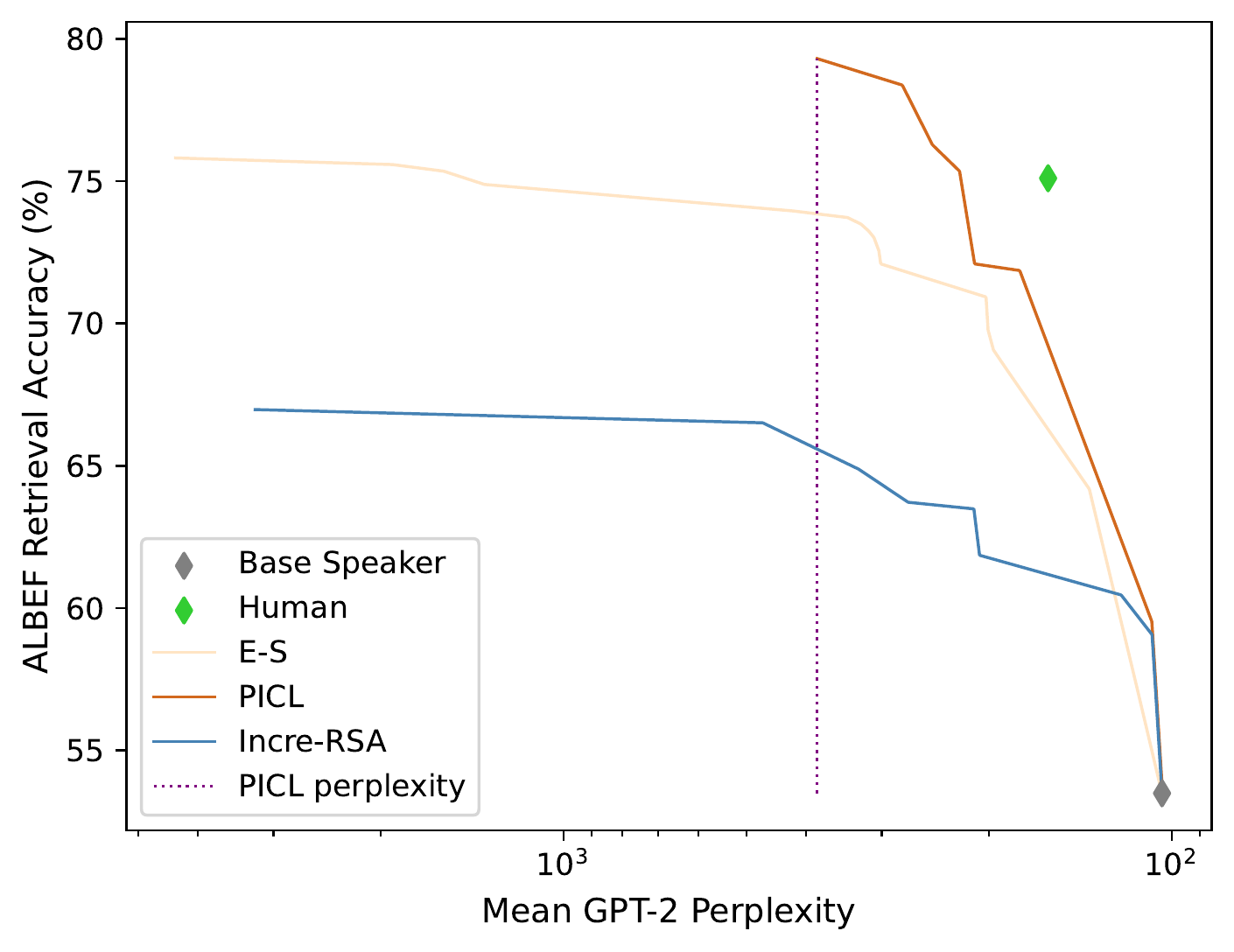}
    \caption{Automatic evaluations show a tradeoff between the informativeness (measured by ALBEF retrieval accuracy) and fluency (GPT-2 perplexity) of discriminative captions on the ImageCoDe valid set in automatic evaluations. Each curve is obtained by varying the value of the informativity hyperparameter.
    Compared with previous methods, our proposed PICL approach achieves a more robust trade-off between fluency and informativeness.
    The vertical line depicts the fluency-controlled criterion (\autoref{subsec:hparam_select}), choosing a perplexity value that matches the perplexity of the maximally-informative PICL. }
    \label{fig:inf_ppl_tradeoff}
\end{figure}

When maximizing predicted evaluative listener accuracy, we observe qualitatively that PICL typically generates captions which are fluent and human-understandable. In contrast, E-S and Incre-RSA are less robust, and under this informativity maximization objective typically produce highly disfluent captions --- identifying captions that are interpretable under our evaluating listener model, ALBEF, but potentially confusing to a human, consistent with past work identifying language drift in reference game setups~\cite{lazaridou-etal-2020-multi,dessi-etal-2022-communication}. This trend is depicted in \autoref{fig:inf_ppl_tradeoff}, where optimizing for high ALBEF accuracy in E-S and Incre-RSA pushes the average GPT-2 perplexity of captions to extremely high values. We will see in human evaluations in \autoref{sec:result} that  the disfluent captions obtained by maximizing predicted informativity in the Incre-RSA and E-S baselines, though ``understandable'' to the ALBEF model, are often uninterpretable for humans. 

\paragraph{Fluency Control}
Given the qualitative failures of E-S and Incre-RSA when maximizing automated proxies for informativity, we propose to improve these baselines using a fluency-controlled optimization scheme that pivots around PICL. In particular, we search for the informativity parameters for E-S and Incre-RSA so that the average GPT-2 perplexity of the generated captions are as close as possible to that of PICL. We refer to the models obtained under this scheme as \textbf{ES (PPL)} and \textbf{Incre-RSA (PPL)}.

\section{Results}
\label{sec:result}
\subsection{Automatic Evaluation}
We use automatic evaluations (\autoref{subsec:auto_eval}) to evaluate the tradeoff between the predicted informativity (using ALBEF) and predicted fluency (using GPT-2) of captions over a wide range of values for the informativity hyper-parameter of each method.

\paragraph{Hyper-parameter Sensitivity}
\autoref{fig:inf_ppl_tradeoff} depicts how each method trades off between discriminativeness and fluency by varying the informativity hyper-parameter. PICL demonstrates higher robustness to hyper-parameter selection than Incre-RSA and ES in the trade-off: while optimizing for ALBEF-predicted informativity-maximization, Incre-RSA and ES produce more corrupted and disfluent captions with high perplexity whereas PICL's perplexity degrades less.

\begin{table}[h]
\centering
\begin{tabular}{lcc}
\toprule
& ALBEF & GPT-2\\
 & \multicolumn{1}{c}{Accuracy} & Perplexity \\ \midrule
Human           & 74.2 & 138.4 \\
Base Speaker    & 54.2 & \textbf{99.4} \\ 
\midrule
\multicolumn{3}{c}{Optimized for Informativity} \\
\midrule
Incre-RSA       & 64.3 & 2703.0  \\
E-S  & \textbf{77.5} & 4093.6 \\
PICL & 77.3 & 380.2 \\
\midrule
\multicolumn{3}{c}{Perplexity-Matched to PICL} \\
\midrule
Incre-RSA (PPL) & 62.9 & 446.5  \\
E-S (PPL) & 73.2 & 366.6 \\ 

\bottomrule
\end{tabular}
\caption{Automatic evaluation results on the ImageCode test set: We evaluate informativity using the retrieval accuracy of the ALBEF evaluative listener using captions generated by each approach. PICL substantially outperforms Base Speaker, Incre-RSA, Incre-RSA (PPL), and E-S (PPL), achieving a competitive level of informativeness to E-S. In fluency, evaluated using GPT-2 perplexity, methods that control for the fluency (PPL) pivoting around PICL achieve similar level of perplexity, while E-S and Incre-RSA that optimized for informativity are substantially less fluent.} 
\vspace{-1em}
\label{tab:acc_test}
\end{table}

\noindent \textbf{Informativeness} \quad As shown in \autoref{tab:acc_test}, PICL substantially outperforms the base speaker and the incremental RSA (Incre-RSA, \citealt{cohn-gordon-etal-2018-pragmatically}) methods on ALBEF retrieval accuracy, and achieves comparable results to emitter-suppressor (E-S, \citealt{Vedantam-etal-2017-ContextAwareCF}). The results demonstrate that our method could leverage CLIP as a listener model in incremental pragmatic caption generation. For both E-S and Incre-RSA, controlling for fluency negatively affects ALBEF accuracy, which conforms with the trend in \autoref{fig:inf_ppl_tradeoff}.

\noindent \textbf{Fluency} \quad 
\autoref{tab:acc_test} also shows the perplexity that GPT-2 assigns to the output of each model on the ImageCoDe test set. As discussed in \autoref{subsec:hparam_select}, Incre-RSA and E-S are less robust when being optimized for informativity, which is reflected by their extremely high perplexity. In contrast, when controlling for the fluency to match PICL's validation perplexity, both Incre-RSA and E-S generate substantially more fluent captions with test perplexity similar to PICL, at the cost of predicted informativeness,  as shown by a drop in ALBEF accuracy.

\subsection{Human Assessment Performance}
We perform human evaluations (\autoref{subsec: human_eval}) to validate these findings about the informativeness and fluency of the discriminative captioning methods.

\begin{table}[t]
\centering

\begin{tabular}{lcc}
\toprule
& Human & Fluency \\
Method & {Accuracy} & Rating \\
\midrule
Human           & \textbf{81.7} & 4.76 \\ 
Base Speaker    & 48.7 & \textbf{4.80}  \\
\midrule
\multicolumn{3}{c}{Optimized for Informativity} \\
\midrule
Incre-RSA & 50.7 & 2.87\\
E-S       & 54.0 & 3.59 \\ 
PICL & \textbf{65.7} & 4.07 \\ 
\midrule
\multicolumn{3}{c}{Perplexity-Matched to PICL} \\
\midrule
Incre-RSA (PPL) & 53.3 & 4.23  \\
E-S (PPL) & 63.7 & 4.54 \\ 

\bottomrule
\end{tabular}
\caption{Human evaluation results on 100 sets of images from ImageCoDe test split: Informativity is assessed by the retrieval accuracy of human annotators using captions generated by each approach. PICL outperforms all other models on human informativeness judgments. For fluency, human annotators evaluate using ratings on a 1-5 scale. Similar to results in the automatic evaluations of fluency (\autoref{tab:acc_test}), annotators assign much lower fluency scores to E-S and Incre-RSA, which do not control for fluency.}
\label{tab:human_acc_test}
\vspace{-1em}
\end{table}

\label{subsec:human_res}
\paragraph{Informativeness} Human retrieval accuracies on model- and human-generated captions are depicted in \autoref{tab:human_acc_test}. In the setting where models are automatically optimized for predicted informativity (\autoref{subsec:hparam_select}), PICL substantially outperforms the Incre-RSA and E-S methods, with gains in human accuracy of 11\% and 15\% respectively. 
The results indicate that captions generated by PICL are more informative than by other approaches, judged by human annotators. 
When we control the disfluency of the other methods to be similar to PICL (as measured by GPT-2 perplexity in automatic evaluations), PICL still substantially outperforms Incre-RSA (PPL) and slightly outperforms ES (PPL). 
Moreover, for both E-S and RSA, controlling for PPL results in more informative captions, which is not reflected in the automatic evaluations using ALBEF (\autoref{tab:acc_test}), implying that disfluency has a more significant negative effect on informativity for humans. While past work has often relied only on automated evaluations, our results indicate that human evaluations are important to accurately compare the performance of discriminative captioning systems.

\paragraph{Fluency} \autoref{tab:human_acc_test} also shows the average fluency scored by human workers for model- and human-generated captions. Similarly to \autoref{tab:acc_test} captions generated by E-S and Incre-RSA without controlling for perplexity are much more disfluent as scored by humans. 

\begin{figure}[t]
\centering
    \includegraphics[width=1.0\linewidth]{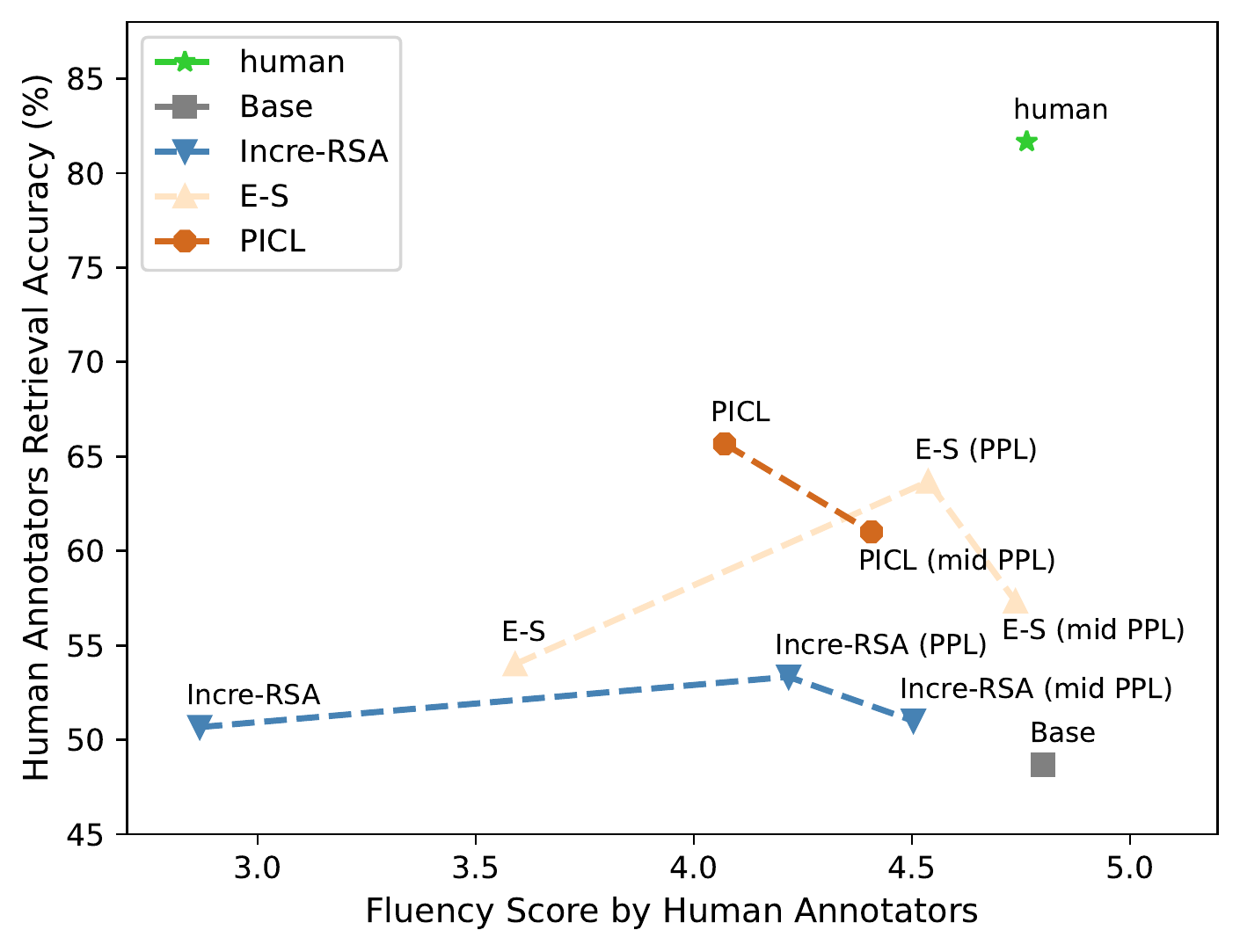}
    \caption{Human eval results on 100 test split static image sets.}
    \label{fig:haf}
    \vspace{-1em}
\end{figure}

\paragraph{Informativity-Fluency Trade-off} 
We further combine the human accuracy and fluency in \autoref{tab:human_acc_test} for each model and plot them in \autoref{fig:haf}. To depict the informativity-fluency trade-off under human assessments, we also include a setting of informativity hyperparameters for each method with an intermediate level of automatically predicted fluency. Specifically, for each model, we search for its informativity parameter so that the average GPT-2 perplexity of generated captions are as close as possible to the average perplexity of the base speaker + PICL. We refer to the models obtained under this scheme as \textbf{ES (mid PPL)}, \textbf{Incre-RSA (mid PPL)} and \textbf{PICL (mid PPL)}.

With the resulting plot shown in \autoref{fig:haf}, PICL outperforms Incre-RSA along both dimensions. In comparison with E-S, PICL achieves better discriminativeness with a loss in fluency. For E-S and Incre-RSA, the trade-off patterns are different from that under ALBEF (\autoref{fig:inf_ppl_tradeoff}). While optimizing for ALBEF accuracy consistently induces more disfluent generation, the optimal informativeness under human judgment is likely to be achieved with a moderate level of disfluency.

\subsection{Automatic vs. Human Evaluation}
The analysis above reflects both agreement and mismatch between automatic evaluation and human judgments on different aspects. To further reveal the correlation between them, and lay a foundation for future work on discriminative captioning to make automatic evaluations more predictive of human performance, we conduct analysis along both axes of informativity and fluency.

\begin{figure}[t]
\centering
    \includegraphics[width=1.0\linewidth]{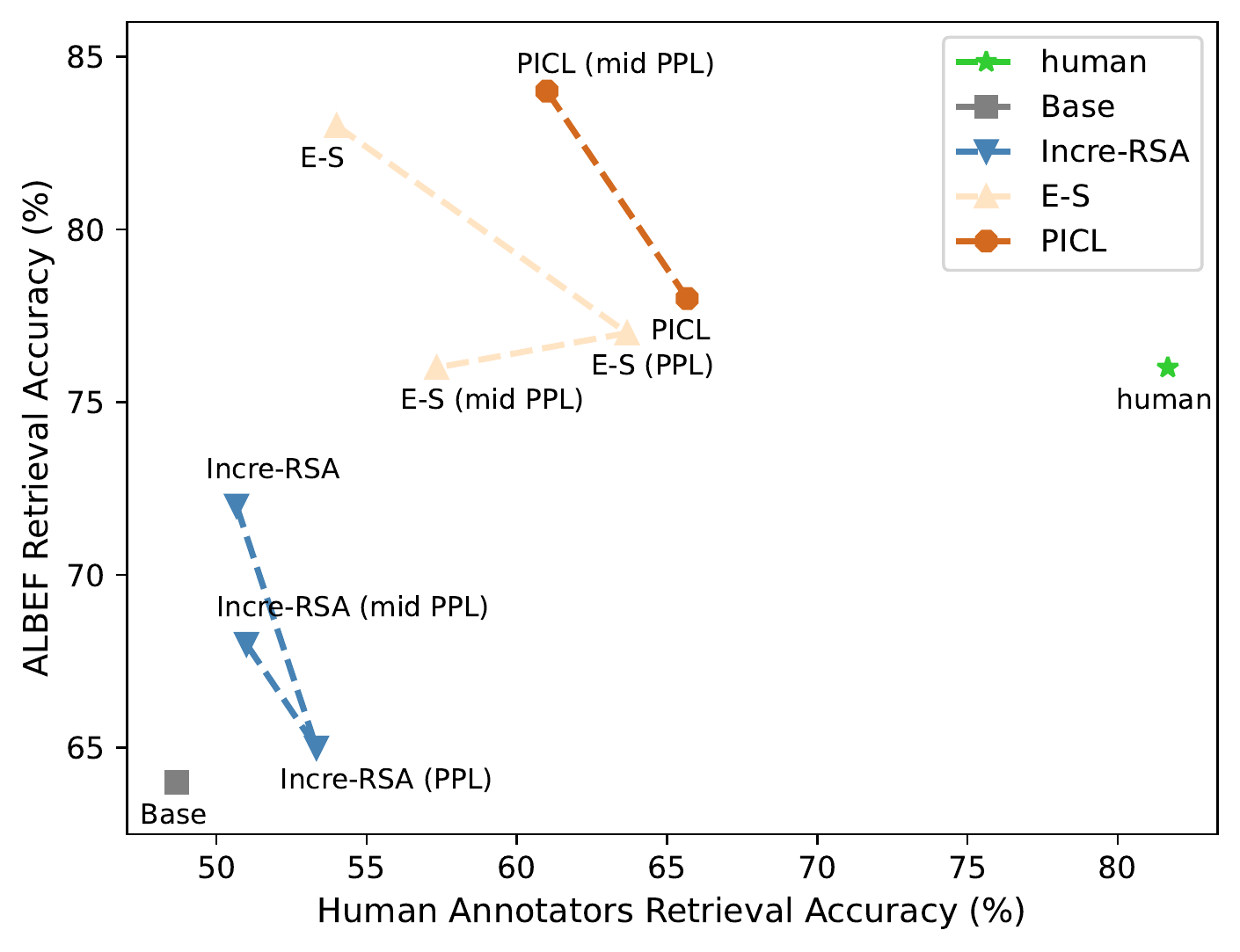}
    \caption{ALBEF accuracy and human accuracy are positively correlated for model-generated outputs, with the exception of disfluent captions produced by the variants of E-S and Incre-RSA that do not control for perplexity.}
    \vspace{-1em}
    \label{fig:albef_vs_human_acc}
\end{figure}

\paragraph{ALBEF vs. Human Retrieval Accuracy}
\autoref{fig:albef_vs_human_acc} plots ALBEF against human retrieval accuracy on the same 100 sets of images. ALBEF accuracy has a strong positive correlation with human judgments except for having human, E-S, and Incre-RSA as outliers. We posit that the performance mismatch on human written captions is because it is challenging for neural retrieval models like ALBEF to interpret human-written descriptions, which are highly nuanced and grammatically complex~\cite{krojer-etal-2022-image}. The high disfluency of the captions of E-S and and Incre-RSA hinders evaluators in interpreting them accurately, despite being discriminative to models. 

\begin{figure}[t]
\centering
    \includegraphics[width=1.0\linewidth]{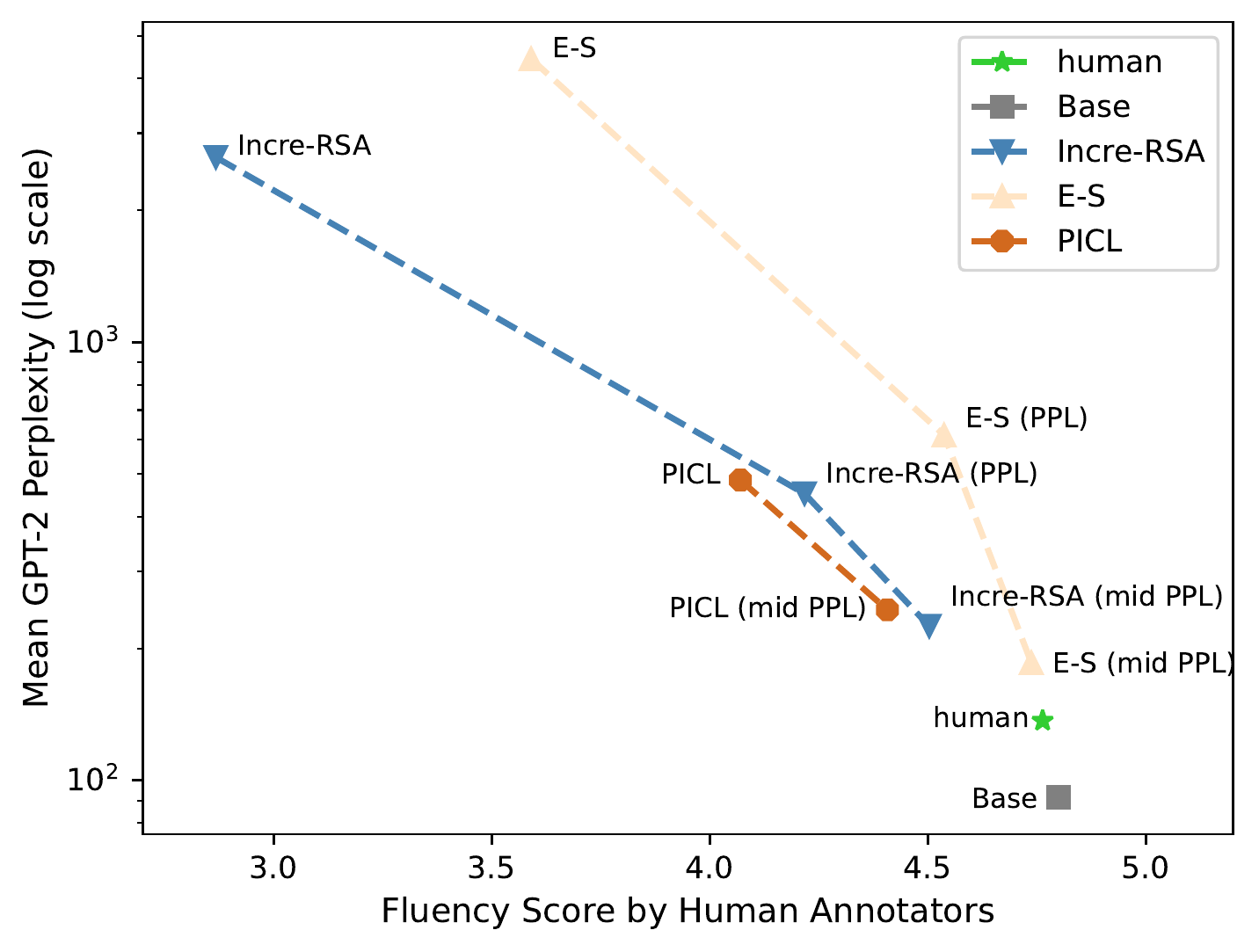}
    \caption{Mean perplexity, under GPT-2, is predictive of human fluency evaluations across systems.}
    \label{fig:ppl_vs_human_fluency}
    \vspace{-1em}
\end{figure}

\paragraph{GPT-2 Perplexity vs. Human Fluency Score}
As illustrated in \autoref{fig:ppl_vs_human_fluency}, on the 100 evaluation image sets, there is a strong correlation between the mean GPT-2 perplexity of captions and human fluency scores, implying that GPT-2 perplexity is a good proxy for human fluency judgments.

\subsection{Ablation Results}
\begin{table}[t]
\centering
\begin{tabular}{lc}
\toprule
  & \multicolumn{1}{c}{Accuracy} \\ \midrule
PICL            & 77.3 \\ \midrule
\quad - incremental & 65.4 \\
\quad - distractors & 57.5 \\

\bottomrule
\end{tabular}
\caption{Automated ablation evaluations of informativeness. 
We evaluate ``- incremental'' that only conducts CLIP scoring and reranking on full captions generated by the speaker model, 
and``- distractor" in which only the target image is included during inference.}
\vspace{-1em}
\label{tab:ablation_test}
\end{table}
To further understand the performance of PICL, we conduct ablation studies to investigate the role of 1) incremental pragmatic inference and 2) grounding language to distinguish from distractors. 

For 1), we experiment with \textbf{PICL - incremental} that removes incremental inference by first using only the base speaker $S_0$ to generate a set of complete and context-agnostic captions, and using CLIP to score these entire captions. For 2), we evaluate \textbf{PICL - distractors}, excluding all distractors and providing only the target image during inference. At each decoding step, the listener distribution is derived by normalizing the CLIP similarities between partial captions and the target image over all candidates. 
As shown in \autoref{tab:ablation_test}, the retrieval accuracy drops substantially on both variations, suggesting that both the incremental inference and grounding to distractors are vital components for pragmatic reasoning in PICL.

\section{Conclusion}
We propose an incremental pragmatic inference approach with a CLIP listener, which combines the strengths of previous approaches that conduct incremental pragmatic reasoning with a separately modeled listener. 
We identify strengths and weaknesses of automatic model-based evaluation of discriminative captioning systems, and suggest that future work 1) control for the disfluency of generated captions and not solely optimize for predicted informativity and 2) use human evaluations.
In human evaluations, our approach outperforms previous discriminative captioning methods, and is substantially more robust than previous approaches in trading off between the fluency and informativity of the captions to human listeners.

\section*{Acknowledgments}
We would like to thank Google for providing funding for this work through a gift on Action, Task, and User Journey modeling, and Samsung Electronics Co., Ltd. for providing funding for BK.

\section*{Limitations}
We evaluate only on the ``static'' image partition of the ImageCoDe dataset. ImageCoDe contains another more challenging partition, containing frames from short temporal intervals in videos, which remains extremely difficult for all current discriminative captioning methods, including our PICL approach. (This partition, along with the static image partition that we use, has previously only been used in contrastive retrieval tasks, not in discriminative captioning.)

While we made a substantial effort to explore the tradeoff between informativity and fluency, we were limited in the number of human evaluations that we were able to do and could only evaluate a few settings of the informativity parameter for each method. We complement these human evaluations with automated evaluations on a much wider range of parameters, and analyze the correlations between human performance and judgements and the automated metrics.

\bibliography{custom,anthology}
\bibliographystyle{acl_natbib}

\appendix
\section{Implementation Details}
\subsection{Computational Resources}
The finetuning of OFA model on COCO captions is run on 4 $\times$ Tesla V100 32GB GPUs. \\

\noindent All pragmatic inference experiments are run on 4 $\times$ GeForce RTX 2080 Ti GPUs.

\subsection{Hyperparameter Searching}
\label{subsec:appendix_hp}
\subsubsection{Rationality Parameters}
\paragraph{Searching Range} The search ranges of the rationality parameter for PICL, E-S, and Incre-RSA are [0, 1], [0, 1], [0, 2] respectively.
\paragraph{Searching Method} We conduct all the hyperparameter searching via coarse-to-fine search, with step sizes 0.1, 0.01, and 0.001 respectively.
\subsubsection{Beam Search Parameters}
For beam search parameters $B, N$ discussed in \autoref{subsec:method_models}, we set $B=16$ and $N=256$.

\subsection{Human Evaluation}
\begin{figure*}[ht]
\centering
    \includegraphics[width=1.0\textwidth]{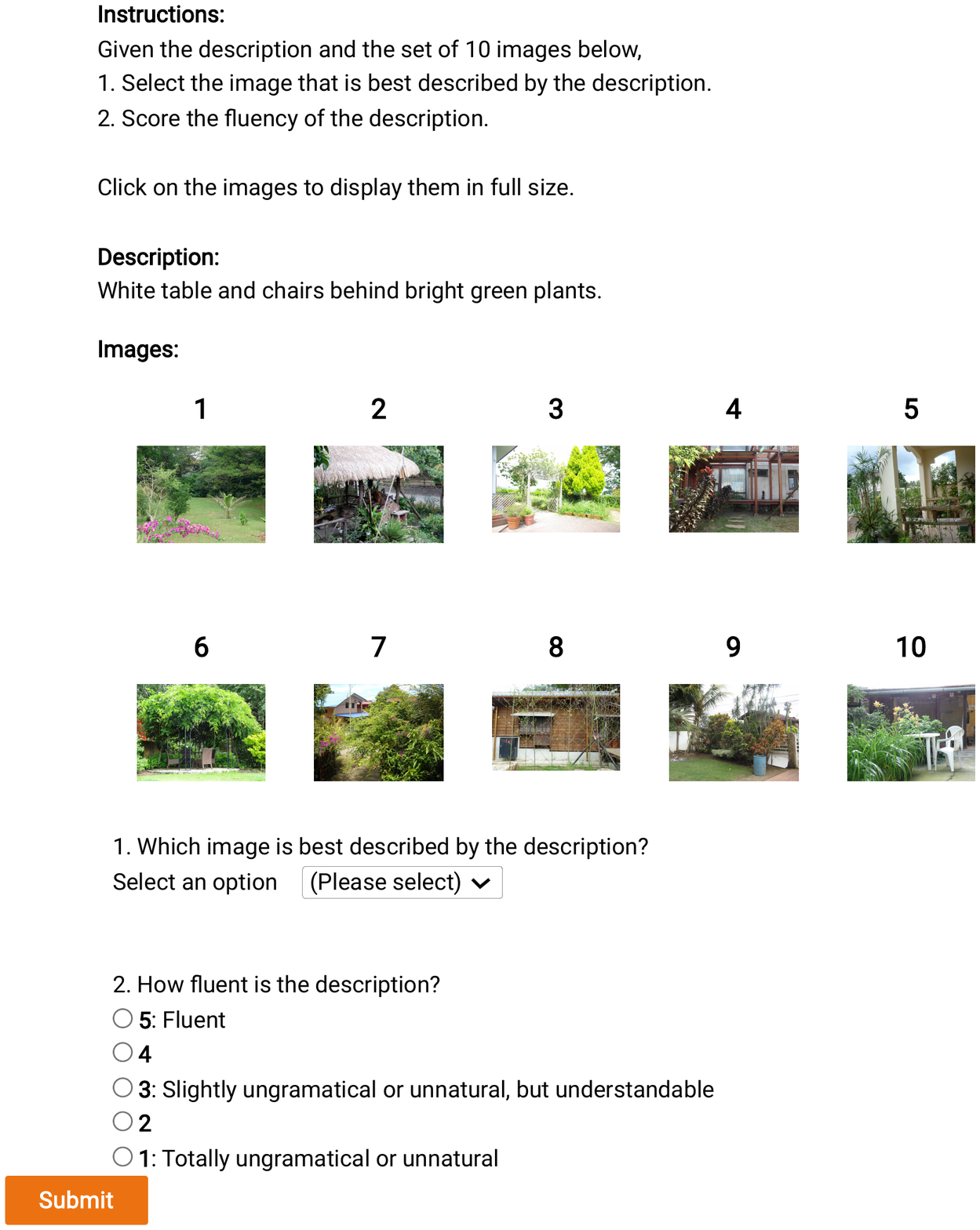}
    \caption{Human Evaluation Interface}
    \label{fig:HIT}
\end{figure*}

\autoref{fig:HIT} shows an example interface of the human evaluation. We have three MTurk workers evaluate each of the 100 instances of (images, caption) for each of the ten configurations of methods (including human-written captions) for informativity (by requiring them to choose the image referred to by the caption) and fluency (on a 1-5 Likert scale) . Workers are paid with \$0.15 per caption evaluation.
\label{subsec:appendix_hit}

\end{document}